\documentclass[sigconf]{acmart}
\AtBeginDocument{%
  }

\setcopyright{acmlicensed}
\copyrightyear{2018}
\acmYear{2018}
\acmDOI{XXXXXXX.XXXXXXX}
\acmConference[Conference acronym 'XX]{Make sure to enter the correct
  conference title from your rights confirmation email}{June 03--05,
  2018}{Woodstock, NY}
\acmISBN{978-1-4503-XXXX-X/2018/06}

\usepackage{multirow}
\usepackage{makecell}
\definecolor{deepgreen}{RGB}{0, 120, 0}  
\definecolor{deepred}{RGB}{198, 14, 14}
\usepackage{graphicx}
\usepackage{bm}

\begin{document}

\title[Into the Unknown: Predicting Individuals' Mobility Beyond Visited Places]{Into the Unknown: Applying Inductive Spatial-Semantic Location Embeddings for Predicting Individuals' Mobility Beyond Visited Places}

\author{Xinglei Wang}
\affiliation{%
  \institution{SpaceTimeLab, UCL}
  \city{London}
  \country{UK}
}
\email{xinglei.wang.21@ucl.ac.uk}
\orcid{0000-0002-9824-7663}

\author{Tao Cheng}
\affiliation{%
  \institution{SpaceTimeLab, UCL}
  \city{London}
  \country{UK}}
\email{tao.cheng@ucl.ac.uk}

\author{Stephen Law}
\affiliation{%
  \institution{Department of Geography, UCL}
  \city{London}
  \country{UK}
}
\email{stephen.law@ucl.ac.uk}

\author{Zichao Zeng}
\affiliation{%
 \institution{3DIMPact \& SpaceTimeLab, UCL}
 \city{London}
 \country{UK}}
\email{zichao.zeng.21@ucl.ac.uk}

\author{Ilya Ilyankou}
\affiliation{%
  \institution{SpaceTimeLab, UCL}
  \city{London}
  \country{UK}}
\email{ilya.ilyankou.23@ucl.ac.uk}

\author{Junyuan Liu}
\affiliation{%
  \institution{SpaceTimeLab, UCL}
  \city{London}
  \country{UK}}
\email{junyuan.liu.22@ucl.ac.uk}

\author{Lu Yin}
\affiliation{%
  \institution{School of Computer Science, University of Surrey}
  \city{Surrey}
  \country{UK}}
\email{l.yin@surrey.ac.uk}

\author{Weiming Huang}
\affiliation{%
  \institution{Institute for Spatial Data Science, University of Leeds}
  \city{Leeds}
  \country{UK}}
\email{W.Huang@leeds.ac.uk}

\author{Natchapon Jongwiriyanurak}
\affiliation{%
  \institution{SpaceTimeLab, UCL}
  \city{London}
  \country{UK}}
\email{natchapon.jongwiriyanurak.20@ucl.ac.uk}

\renewcommand{\shortauthors}{Wang et al.}

\begin{abstract}

Predicting individuals' next locations is a core task in human mobility modelling, with wide-ranging implications for urban planning, transportation, public policy and personalised mobility services. Traditional approaches largely depend on location embeddings learned from historical mobility patterns, limiting their ability to encode explicit spatial information, integrate rich urban semantic context, and accommodate previously unseen locations. To address these challenges, we explore the application of CaLLiPer\textemdash a multimodal representation learning framework that fuses spatial coordinates and semantic features of points of interest through contrastive learning\textemdash for location embedding in individual mobility prediction. CaLLiPer’s embeddings are spatially explicit, semantically enriched, and inductive by design, enabling robust prediction performance even in scenarios involving emerging locations. Through extensive experiments on four public mobility datasets under both conventional and inductive settings, we demonstrate that CaLLiPer consistently outperforms strong baselines, particularly excelling in inductive scenarios. Our findings highlight the potential of multimodal, inductive location embeddings to advance the capabilities of human mobility prediction systems. We also release the code and data (https://github.com/xlwang233/Into-the-Unknown) to foster reproducibility and future research .

\end{abstract}



\received{20 February 2007}
\received[revised]{12 March 2009}
\received[accepted]{5 June 2009}

\maketitle

\section{Introduction}
\label{sec:intro}
Human mobility—the movement of individuals across space and time—is a fundamental aspect of human behaviour, shaping urban dynamics, transportation systems, and public policies \cite{barbosa2018human}. Over the past decade, computational approaches, especially those based on artificial intelligence (AI) techniques such as deep learning, have played an increasingly prominent role in human mobility modelling. Advancing these computational approaches has become a major future direction for human mobility science \cite{pappalardo2023future}. Within this research field, predicting individual’s next location is a key task \cite{luca2021survey}, for which numerous models and algorithms have been proposed. Despite varying architectures, almost all of these models operate on proper location representations. The most widely adopted strategy is to encode locations as dense vector embeddings, which serve as compact and informative inputs for downstream mobility prediction models \cite{lin2021pre}. However, existing location embedding pre-training methods for individual-level next location prediction typically rely on historical mobility data to learn the co-occurrence or sequential patterns of visited places. This results in several limitations.

First, the resulting location representations do not explicitly encode spatial information (e.g., geographical coordinates), despite it being universally acknowledged as a crucial factor influencing individual’s travel. This negligence could limit the downstream prediction model’s ability to account for the spatial dependencies in mobility behaviour.

Second, the pre-trained embeddings are tied to a fixed set of locations. As new mobility data are collected and fed into the location prediction system, new locations might emerge, either from new users or from existing users who begin visiting previously unseen locations (this is a common behavioural change). Existing models cannot accommodate these “new locations” and thus the downstream prediction performance would be limited. Although one could re-train the whole location embedding model to incorporate these new locations, it incurs extra compute, which is not ideal for real-world prediction systems that favour fast responses. 

Moreover, human mobility is also heavily influenced by the urban environment, such as transportation networks, terrain, and land use \cite{buchin2012context}. Among these spatial contextual features, land use - reflecting urban functions- is particularly relevant \cite{lee2015relating}. Although studies have shown that capturing the semantic characteristics of locations through points of interest (POIs) can enhance individual mobility prediction \cite{hong2023context}, existing location embedding approaches fail to effectively incorporate this information. 

We argue that addressing these limitations requires location representations that are spatially explicit, semantically informed, and inductive – i.e., capable of generalising to unseen locations. To this end, we propose to apply CaLLiPer \cite{wang2025multi}, a recently developed representation learning method originally designed for learning urban space representations from POIs, to represent locations in human mobility modelling domain. CaLLiPer enables multimodal representation learning by combining spatial (coordinate) and platial (POIs semantic) information through contrastive learning. It employs a location encoder to capture spatial information and aligns the resulting location embeddings with corresponding POI representations generated by a text encoder. As a result, CaLLiPer produces embeddings that are spatially explicit, semantically rich, and inductive by design. It has also been proved in the original paper that, being a multimodal location embedder, CaLLiPer performs better than other state-of-the-art methods in characterising fine-scale urban spaces.

To evaluate the effectiveness of utilising CaLLiPer for location representation in mobility prediction, we conduct extensive experiments on four public human mobility datasets under two distinct settings, i.e., \textit{conventional} and \textit{inductive}. The conventional setting aligns with what has been commonly adopted in existing studies, while the inductive setting simulates real-world scenarios where new locations emerge as new users enter the system or existing users alter their routines and visit unfamiliar places. The results show that, in two out of the four chosen datasets, CaLLiPer significantly outperforms competitive baseline models across all evaluation metrics, with a particularly notable advantage under the inductive condition. In the remaining two datasets, although it does not achieve the best performance across all metrics in both settings, the application of CaLLiPer embeddings still produces strong results – either outperforming baselines under the inductive setting or achieving top performance on a subset of the metrics consistently. These experimental results demonstrate the practical value of utilising CaLLiPer-generated embeddings for individuals' mobility prediction, especially in inductive scenarios. 

We summarise our contributions as follows: 
\begin{itemize}
    \item Novel application: We are the first to apply multimodal location encoding—characterised by spatial explicitness, semantic awareness, and inductive capability—to location embedding for individual next location prediction. 
    \item Addressing a practical issue: Our approach tackles the real-world challenge of handling emerging (unseen) locations in next location prediction systems. 
    \item Empirically validated and reproducible: We conduct extensive experiments across conventional and inductive settings to demonstrate the effectiveness of the application of CaLLiPer. The code and data are made publicly available to support reproducibility and provide a benchmark for the research community.
\end{itemize}

The remainder of this paper is structured as follows: Section \ref{sec:related_work} reviews related work in individual mobility prediction and location encoding, and explains our rationale for choosing CaLLiPer as the applied method. Section \ref{sec:method} introduces our methodology, covering notations, the problem statement, the methodological framework, and preliminaries concerning the CaLLiPer model and downstream prediction models. Section \ref{sec:exp} details our experimental setup, with a focus on the implementation of conventional and inductive settings. Section \ref{sec:res} presents our quantitative and qualitative empirical results. In Section \ref{sec:discussion}, we analyse the fundamental differences between the CaLLiPer-based location embedding method and other approaches, and discuss the implications of our empirical findings. Finally, Section \ref{sec:conclusion} summarises the study and outlines future directions.

\section{Related Work}
\label{sec:related_work}
\subsection{Location Embedding for Next Location Prediction}

Most next location prediction models require locations to be represented by latent embedding vectors. The most straightforward strategy is to use an embedding layer, which is essentially a lookup table that stores embedding vectors for a location set of fixed size \cite{kong2018hst,zhao2020go}. This embedding layer is typically trained end-to-end with task-specific objectives. However, such embeddings are difficult to transfer to other models and tasks. They are also prone to overfitting problems and struggle to incorporate comprehensive information about locations, such as their semantic meaning or functional use of locations \cite{lin2021pre}. 

To tackle these issues, researchers have proposed to pre-train location embeddings using unsupervised or self-supervised objectives to incorporate more general and comprehensive information of locations. Inspired by distributed word representations widely used in natural language processing (NLP) domain \cite{mikolov2013distributed}, researchers have proposed to treat locations in people’s mobility series as words in sentences, applying Word2Vec models \cite{mikolov2013efficient} to capture the co-occurrence patterns of locations \cite{yao2018representing,zhou2018deepmove}. Subsequent methods further adapted Word2Vec models to integrate additional information into the embeddings. For example, special binary tree structures were constructed to account for the spatial proximity of locations \cite{feng2017poi2vec} or the time locations are visited \cite{wan2021pre}, and used in the hierarchical Softmax calculation in CBOW model \cite{mikolov2013efficient}. More recently, one work leveraged the BERT \cite{devlin2019bert} architecture to derive dynamic latent embeddings for locations based on their contextual neighbours \cite{lin2021pre} and achieved state-of-the-art performance on two mobile phone signalling datasets. Despite these advancements, existing methods share several common limitations. 

With the exception of a few methods like POI2Vec \cite{feng2017poi2vec}, most approaches largely ignore the spatial attributes of locations (such as geographic coordinates), even though such spatial attributes depicting distance and proximity actually plays a crucial role in people's travel. Moreover, they fail to consider the semantic context of urban places, which is another important factor shaping human mobility behaviour. 

The most prominent issue is their reliance on static, pre-defined location sets. These sets are typically derived from the mobility data itself, and the process varies depending on the dataset type. For general GNSS tracking datasets, this involves identifying individual stay points, followed by the spatial clustering of stay points across all users to define the final set of locations \cite{hong2023context}. In POI check-in datasets generated by location-based social network (LBSN) applications, the locations correspond to the POIs contained within the dataset. For existing methods, the delineation of locations\textemdash both their number and spatial whereabouts\textemdash is fixed during pre-training and remains unchanged at inference, making it difficult to accommodate scenarios where new locations emerge over time. Addressing this limitation is the primary goal of this paper.

\begin{figure*}[h]
  \centering
  \includegraphics[width=0.75\linewidth]{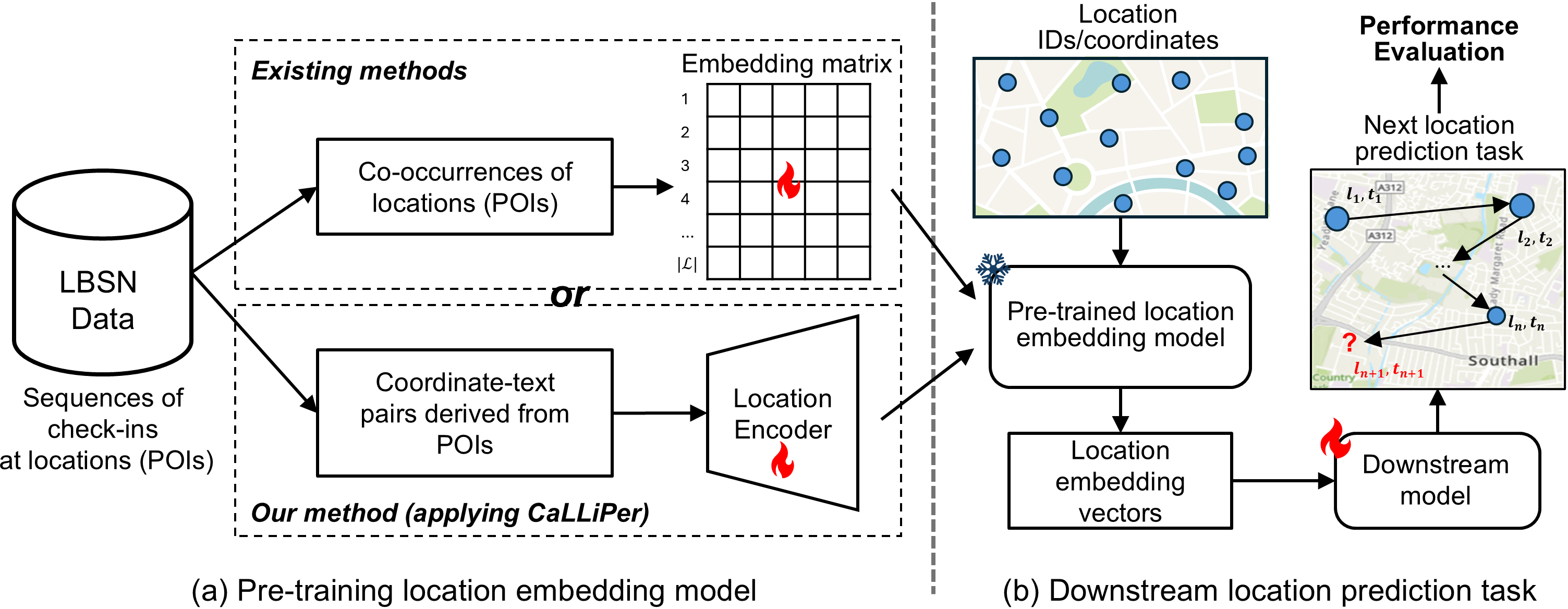}
  \caption{The methodological framework, consisting of \textit{pre-training} and \textit{downstream prediction} stages. The focus is on the pre-training, where we experiment both existing methods and our proposed method (i.e., applying CaLLiPer). The effectiveness of different methods are indicated by the performance of the downstream location prediction model.}
  \label{fig:framework_overview}
\end{figure*}

\subsection{Location Encoding}
\label{sec:loc_encoding}

Location encoding refers to the process of embedding point locations into a vector space so that these location embeddings can be readily used in downstream neural network modules \citep{mai2022review}, with various aims like geographic prior modelling \citep{chu2019geo,klemmer2025satclip,mac2019presence,wang2025multi} or spatial context modelling \citep{mai2020iclr}, etc.

The motivation for location encoding was first comprehensively articulated in \citep{mai2022review}, in which the authors provided a general conceptual framework that unifies the formulation of location encoding methods. Following their formulation, location encoding methods generally take the form of $y=\text{NN}(\text{PE}(\lambda,\phi))$, where a geographical or projected coordinate $(\lambda,\phi)$ is processed through a parametric positional encoding (PE) function and a neural network (NN). 

The NN component is usually implemented as a fully connected residual network (FC-Net) or sinusoidal representation network (SirenNet)  \cite{russwurm2024sh}, while PE methods vary, including Wrap \citep{mac2019presence}, Grid and Theory \citep{mai2020iclr}, Sphere* \citep{mai2023sphere2vec}, and Spherical Harmonics (SH) \citep{russwurm2024sh}, etc. Depending on the chosen PE function, location encoding can operate over either planar or spherical geometries at different spatial scales.

Apart from the architecture design of PE and NN, the training of NN is also central to location encoding. Depending on the target features, different training methods have been used. For simple binary classification tasks (e.g., distinguishing land vs. ocean), supervised learning with binary cross-entropy loss is typically used \cite{russwurm2024sh}. For more complex objectives, such as aligning locations with images \cite{klemmer2025satclip} or textual descriptions \cite{wang2025multi}, contrastive learning is employed to integrate imagery or textual modalities into the location encoder. 

Theoretically, location encoding enables continuous embedding of geographic space, allowing for vector representations at every possible point location. This makes it highly inductive\textemdash capable of generalising to unseen locations during inference \cite{mai2022review,wang2025multi}.

Location encoding has been applied in a variety of domains, including geo-aware image classification \citep{mac2019presence}, POI classification \citep{mai2020iclr}, land use classification and socioeconomic status distribution mapping \cite{wang2025multi}, etc. However, to our knowledge, it has not yet been applied to generate location embeddings for human mobility tasks\textemdash an application gap addressed by this paper.

As discussed in the introduction, effective location prediction requires embeddings that are spatially explicit, semantically rich, and inductive. While several location encoding methods meet these criteria, in this study we chose to apply CaLLiPer over other alternatives like SatCLIP \cite{klemmer2025satclip}. The rationale is that POI data is more suitable for this task. Our justification is threefold:

First, POI data offer unique advantages. Visual features can be ambiguous; for example, two urban areas/streets may look similar but serve different functions. Whereas POI data are more assertive and precise in their semantic meanings\textemdash a shopping mall cannot be mistaken for an office building after all. POI-based descriptions also better capture nuanced, people-centric semantics \cite{wang2025multi,ilyankou2023supermarket}. Moreover, the original CaLLiPer paper \cite{wang2025multi} has demonstrated its superior performance over other models like Space2Vec \cite{mai2020iclr} and SatCLIP \cite{klemmer2025satclip} in characterising fine-scale urban spaces\textemdash essentially equivalent to "locations" in urban human mobility context.

Second, spatial resolution and coverage are important considerations. While satellite imagery is useful for constructing location-image pairs at a global scale, it is less effective for fine-grained, local-scale training. It is unlikely for two nearby locations, only metres apart, to have distinct, non-overlapping satellite images. Street-view imagery, while useful for capturing more nuance urban characteristics, often suffers from spatial distribution bias, especially in open-source datasets, leading to inconsistent coverage across urban areas \cite{fan2025coverage}.

Third, major sources of human mobility data, such as LBSNs, already include detailed POI information. This makes POI data a natural and efficient input for models like CaLLiPer.

\section{Methodology}
\label{sec:method}
In this section, we introduce the concepts and notations used in this article, formulate the research problem of pre-training location embeddings, explain the methodological framework, and provide a brief introduction to CaLLiPer and the downstream prediction model.

\subsection{Notations and Problem Statement}

\paragraph{Definition 1 – Individual’s mobility} An individual’s movements during a certain period can be represented by a spatio-temporal trajectory $s$ consisting of sequential visiting records. A visiting record $(u, L, t)$ indicates that user $u$ visited location $L$ at time $t$. We denote the set of individuals’ trajectories as $\mathcal{S}$, the set of all locations appear in the dataset as $\mathcal{L}$, and the set of all individuals as $\mathcal{U}$.

\textit{Definition 2 - Location}. A location $L$ can generally be represented as $L=(l, c, g)$, where $l$ is the location identifier, $c$ encodes its context semantics such as the surrounding land use, and g denotes the geometry of the location. The delineation of locations, i.e., $g$, varies depending on the type of mobility data. For a GNSS tracking dataset, each location is typically defined as the convex hull of a set of spatially proximate stay points derived from user’s trajectories, whereas for a location-based social network (LBSN) check-in dataset, each location corresponds to the respective POI in those LBSN applications. 

\textit{Problem Statement –Pre-training location embedding models (for mobility prediction)}. The aim is to pre-train a parameterised mapping function $\mathcal{F}$ to generate an embedding vector for a target location $L$ based on the available data, e.g., individuals’ trajectories and/or POI data. The effectiveness of the pre-trained location embedding models will be verified by a mobility prediction task. 

\subsection{Methodological framework}
\label{sec:method_framework}

Figure \ref{fig:framework_overview} presents the methodological framework of this study, demonstrating the complete workflow of the pre-training and downstream application stages.

The objective is not to invent new models, but rather to apply existing ones to address practical issues identified in real-world applications. Pre-training location representations based on the co-occurrence of locations manifested in mobility data is argued to be sub-optimal, particularly as this approach struggles to accommodate new locations. Instead, we hypothesise that CaLLiPer — which pre-trains location representations using general spatial (coordinates) and semantic (textual descriptions) information — is more suitable for location prediction tasks, especially in handling previously unseen locations. To verify whether the hypothesis holds, both the baseline methods and the proposed approach are implemented. The resulting pre-trained location embedding models (whether they are embedding matrices or location encoders) are frozen and subsequently integrated into downstream prediction models to perform the next-location prediction task. The performance of the downstream models, enhanced with the pre-trained location embeddings, serves as an indicator of the effectiveness of the embedding models. 

It is important to note that Figure 1 uses LBSN data as the primary data source. LBSN data alone is sufficient to complete the entire process, as it contains both individuals’ mobility information and POIs. However, for other types of datasets that may not include POIs, it becomes necessary to incorporate additional POI data from external sources to facilitate the pre-training of CaLLiPer. This scenario is addressed in the experiments, as detailed in Section 4.1, where Foursquare POI data is integrated to facilitate experiments on the Gowalla-LD and Geolife datasets.

\subsection{Preliminaries about CaLLiPer and Downstream Prediction Model}

\subsubsection{CaLLiPer}

\begin{figure}[h]
  \centering
  \includegraphics[width=0.98\linewidth]{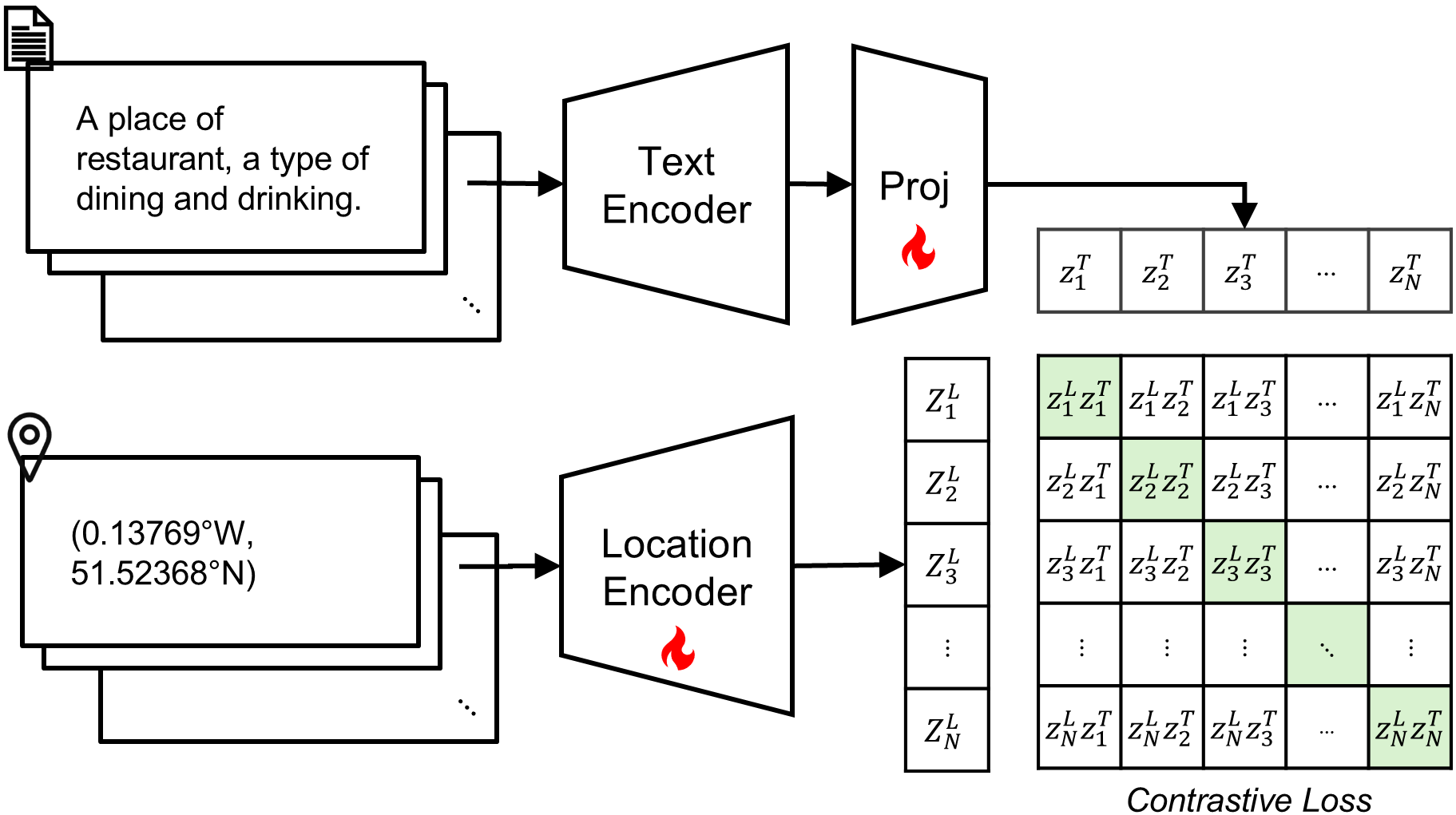}
  \caption{The model framework of CaLLiPer.}
  \Description{A woman and a girl in white dresses sit in an open car.}
  \label{fig:calliper}
\end{figure}

As our study is centred around the application of the recently proposed urban space representation learning model - CaLLiPer, we provide a brief introduction to its overall architecture. As Figure \ref{fig:calliper} shows, CaLLiPer consists of three main components: (1) a location encoder $f^L$ that embeds individual coordinates into higher dimensional space (further details can be found in the Appendix \ref{append:calliper}); (2) a text encoder $f^T$ for extracting semantic features from POI’s natural language descriptions; and (3) a projection layer $f^P$ that projects the output of text encoder to match the dimensions of the location embeddings. 

In the pre-training stage, the location encoder embeds a batch of $N$ POI coordinates $\boldsymbol{x}^L \in \mathbb{R}^{N \times 2}$ into location embeddings $\boldsymbol{z}^L \in \mathbb{R}^{N \times d}$ in a $d$-dimensional space. Simultaneously, the corresponding POI textual descriptions $x^T \in \mathbb{R}^{N \times t}$ ($N$ sentences with arbitrary length $t$) pass through the text encoder and projection layer and be mapped to text embeddings $z^T \in \mathbb{R}^{N \times d}$. The mathematical formulations are as follows:

\begin{equation}
    f_{\Theta ^L}^L (x^L ) = z^L \in \mathbb{R}^{N \times d}
\end{equation}

\begin{equation}
    f_{\Theta ^P}^P (f^T (x^T))= z^T \in \mathbb{R}^{N \times d}  
\end{equation}

The parameters of the location encoder and the projection layer, i.e., $\Theta^L$ and $\Theta^P$, are optimised, while the text encoder is frozen and do not receive parameter update. The simple yet highly effective bi-directional InfoNCE is adopted as the training objective \citep{radford2021clip}:

\begin{align}
Obj(\Theta^L,\Theta^P) = & -\frac{1}{2N} \bigg[ 
\sum_{i=1}^{N} \log \frac{ \exp(z_i^L z_i^T / \tau) }{ \sum_{j=1}^{N} \exp(z_i^L z_j^T / \tau) } \nonumber \\
& + \sum_{i=1}^{N} \log \frac{ \exp(z_i^T z_i^L / \tau) }{ \sum_{j=1}^{N} \exp(z_i^T z_j^L / \tau) } 
\bigg]
\end{align} where $\tau$ is a temperature hyperparameter.

Once the pre-training is finished, the location encoder can be used in various downstream tasks in a “plug and play” manner, producing the embeddings for any urban spaces without further training, while an additional downstream model (also referred to as downstream predictor) is optimised to obtain the final target features. 

\subsubsection{Downstream Model}

The embedding vectors generated by different location embedding methods constitute the input to a downstream prediction model to predict individuals’ next location. The downstream location prediction model is typically based on sequence models, e.g., LSTM \cite{hochreiter1997long} or Transformer \cite{vaswani2017attention}. Next location prediction task has traditionally been formulated as a multi-class classification problem, where the final output is a probability distribution $P(\hat{l}_{n+1})$ depicting the probability of the user visiting each location at the next time step $n+1$. Therefore, no matter what the architectures of the sequence models are, the last layer would commonly be a fully-connected (FC) layer followed by a softmax function:

\begin{equation}
\label{eq:pred_prob}
    P(\hat{l}_{n+1}) = Softmax(f^{FC}(h_n))
\end{equation}
where $f^{FC}$ denotes the FC layer and $h_n$ denotes the hidden state at the $n$th time step.

The trainable parameters of the downstream model are optimised by the multi-class cross-entropy loss (CEL):

\begin{equation}
    CEL = -\sum_{k=1}^{|\mathcal{L}|}P(l_{n+1})^{(k)}log(P(\hat{l}_{n+1})^{(k)})
\end{equation}
where $P(\hat{l}_{n+1})^{(k)}$ denotes the predicted probability of visiting the $k$th location and $P(l_{n+1})^{(k)}$ is the one-hot vector representing the ground truth. 

In this paper, we employ MHSA \cite{hong2023context} as the downstream model, for it is recently proposed, well-documented, open-sourced and based on Transformer architecture, which has been a common backbone adopted in next location prediction models.

\section{Experiment}
\label{sec:exp}

We conduct extensive experiments, following the framework introduced in Section \ref{sec:method_framework}. In particular, we incorporate the location embeddings generated by different models, including competitive baselines and CaLLiPer, into a common downstream next location prediction model and compare their results to evaluate the effect of CaLLiPer as a location embedding method. 

\subsection{Data and Preprocessing}

\paragraph{\textbf{Mobility data}} We use three commonly adopted, publicly available datasets in our experiments: two location-based social network (LBSN) check-in datasets—Foursquare check-ins in New York City (denoted as FSQ-NYC) \cite{yang2014modeling}, and the Gowalla dataset restricted to London, UK (denoted as Gowalla-LD) \cite{cho2011friendship}, as well as one GNSS tracking dataset, Geolife \cite{zheng2010geolife}.

\paragraph{\textbf{POI data}} POI data are required for training CaLLiPer. Since FSQ-NYC and FSQ-TKY already include POI information, no additional sourcing is needed. For Gowalla and Geolife, which lack detailed POI data, we obtained the necessary data for London and Beijing from the Foursquare Open Place dataset. While alternative sources such as Overture and OSM are available, we chose Foursquare’s Open-Source Places dataset \cite{fsq2024osp} due to its detailed classification scheme, business-grade coverage and quality, and ease of access via its APIs.

\paragraph{\textbf{Data preprocessing}} We follow the common preprocessing practices introduced in \cite{hong2023context}. For check-in data, we excluded unpopular POIs with fewer than 10 check-ins and filtered out users with fewer than 10 records. For Geolife, the preprocessing procedures include filtering users with too few data points (less than 50 days), identifying stay points, and applying spatial clustering to derive locations. Trackintel library \cite{martin2023trackintel} was used in this process.

The basic statistics of the resulting dataset after preprocessing are presented in Table \ref{tab:data_stats}.

\begin{table*}
  \caption{Basic statistics of the mobility datasets after preprocessing. The mean and standard deviation across users are reported.}
  \label{tab:data_stats}
  \begin{tabular}{lcccc}
    \toprule
     &FSQ-NYC&FSQ-TKY&Gowalla-LD&Geolife\\
    \midrule
    $\#$Users & 535& 1643& 68& 39\\
    \#Days tracked & 319 & 318& 367& 1703\\
    \#Total unique locations&	4019&	6959&	1065& 1248\\
    \#Stays per user & $178.3\pm206.1$ &	$225.1 \pm 220.9$&	$127.1 \pm 123.3$&	$358.6 \pm 383.6$\\
    \#Stays per user per day& $2.1 \pm 2.1$& $2.7 \pm 2.7$ &	$2.4 \pm 3.0$ &	$2.4 \pm 1.5$\\
    \#Unique locations per user&	$34.7 \pm 21.4$&	$56.6 \pm 33.5$&	$46.3 \pm 44.2$&	$61.5 \pm 52.8$\\
  \bottomrule
\end{tabular}
\end{table*}


\subsection{Baseline Methods}


The following location embedding methods are selected as baselines in the experiment. This is a fairly comprehensive and representative list of models, from basic (Vanilla-E2E) to state-of-the-art.

\begin{itemize}
    \item Vanilla-E2E: A vanilla embedding layer trained end-to-end (E2E), which is essentially a lookup table for a fixed set of locations, as commonly used in prior work \cite{kong2018hst,zhao2020go}.
    \item Skip-gram \cite{mikolov2013efficient}: A Word2Vec model that has been utilised for modelling mobility trajectories \cite{liu2016exploring}.
    \item POI2Vec \cite{feng2017poi2vec}: A Word2Vec-based method, which models spatial information through assigning locations into a geographical binary tree. 
    \item Geo-Teaser \cite{zhao2017geo}: Geo-Temporal Sequential Embedding Rank model, which integrates temporal and spatial features through vector expansion and a modified negative sampling strategy.
    \item TALE \cite{wan2021pre}: Time-Aware Location Embedding that incorporates temporal information through designing a temporal tree structure for hierarchical softmax calculation.
    \item CTLE \cite{lin2021pre}: A context and time-aware method based on BERT, which generates location embeddings considering neighbouring context in trajectories.
\end{itemize}


\subsection{Conventional and Inductive Settings}

\begin{figure}[h]
  \centering
  \includegraphics[width=0.7\linewidth]{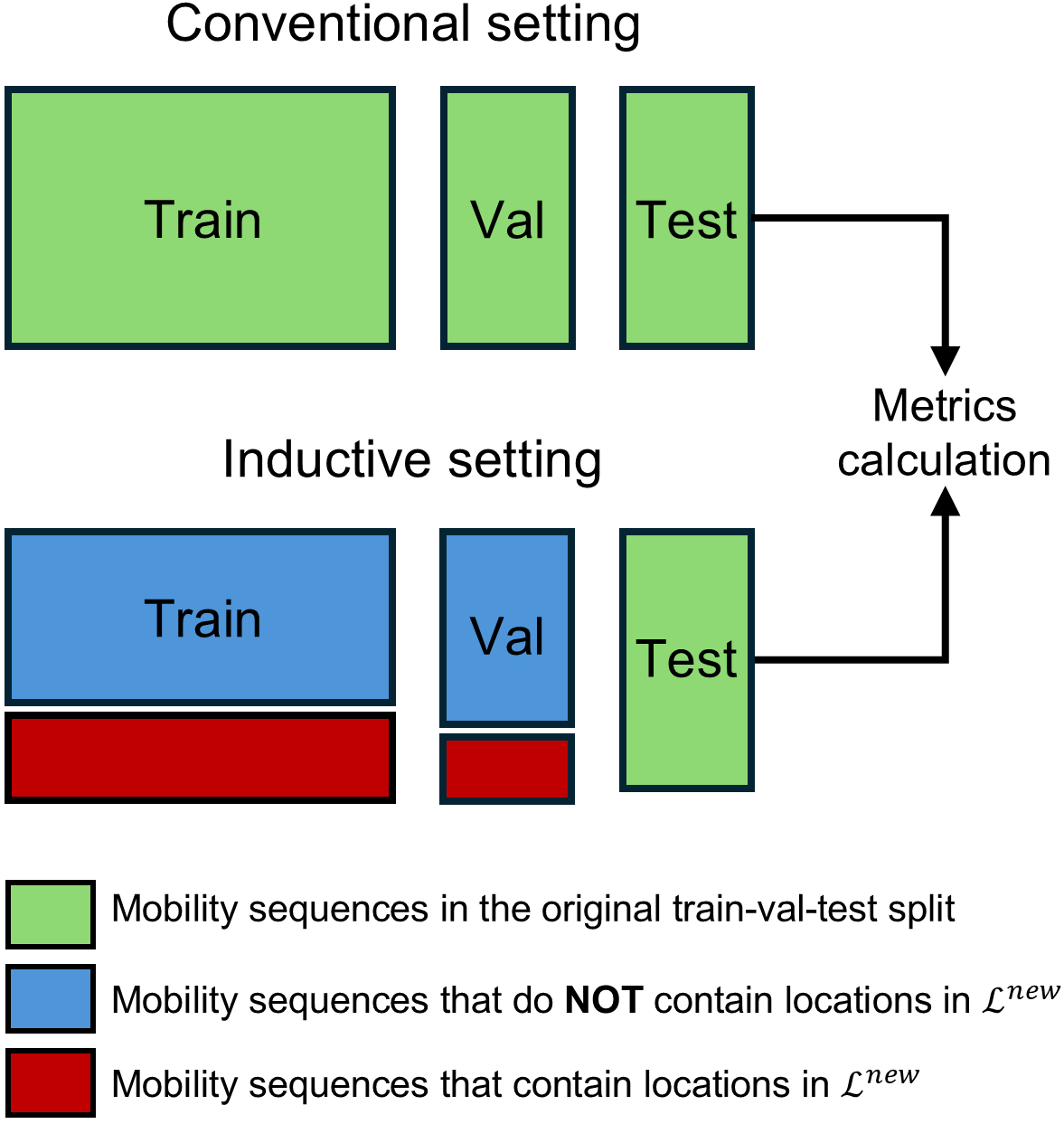}
  \caption{The illustration of the process of constructing train, validation and test sets under two setting.}
  \Description{A woman and a girl in white dresses sit in an open car.}
  \label{fig:exp_settings}
\end{figure}

The key implementational differences of the conventional and inductive settings lie in the construction of the train, validation, and test sets. Figure \ref{fig:exp_settings} illustrates how these sets are constructed for the model training and evaluation under the two experimental settings.

\textbf{\textit{Conventional setting}}. All datasets are split into non-overlapping train, validation, and test sets with a 6:2:2 ratio based on time. Mobility sequences (i.e., users’ location visit sequences) are created using a sliding window of seven days, following empirical findings from \cite{hong2023context}, which show that models perform best when using the past seven days of data. As such, sequences from the first 60\% of tracking days are used for training, and the last 20\% for testing. For each model, we run the experiment five times and report the mean and standard deviation for each evaluation metric (shown in the left part of Table \ref{tab:perf_res}).

\textbf{\textit{Inductive setting}}. To simulate unseen locations, we modify the conventional data splits. Specifically, we randomly sample 10\% of the locations in the training set and denote this subset as $\mathcal{L}^{new}$. Then we remove all the mobility sequences that contain any locations in $\mathcal{L}^{new}$ from the original training and validation sets to form their inductive counterparts. The test set remains unchanged, which means that it contains certain locations that are not seen during neither the pre-training phase nor the downstream model training phase. This setup allows us to evaluate how well different location embedding models generalise to new locations. 

To account for sampling variability, we repeat this process with five different random samples of $\mathcal{L}^{new}$, resulting in five different training and validation sets. For each model, we run five experiments, each corresponding to one of these five data splits. We then report the mean and standard deviation of the evaluation metrics across these five experiments (shown in the right part of Table \ref{tab:perf_res}).

\subsection{Evaluation Metrics}

We adopted the following commonly used metrics to quantify the predictive performance of compared models.

\textbf{\textit{Accuracy}}. Predictions are sorted in descending order based on their probability of being the next location, and Acc@k measures the proportion of times the ground truth location appears within the top-k predictions. In location prediction literature, this metric is also referred to as Recall@k or Hit Ratio@k. In our experiment, Acc@1, Acc@5, and Acc@10 were reported for evaluation.

\textbf{\textit{Mean reciprocal rank (MRR)}}. This metric calculates the average rank reciprocal at which the first relevant entry was retrieved in the prediction vector:
\begin{equation}
    MRR = \frac{1}{N} \sum_{i=1}^N{\frac{1}{rank_i}}
\end{equation}
where $N$ denotes the number of test samples and $rank_i$ is the rank of the ground truth location in $P(\hat{l}_{n+1})$ (the probability distribution predicted by the downstream model; see Equation \ref{eq:pred_prob}) for the i$th$ test sample.

\textbf{\textit{nDCG@k}}. Normalised discounted cumulative gain (with rank position $k$) measures the ranking quality of a prediction vector by the ratio between the discounted cumulative gain (DCG) and the ideal discounted cumulative gain (IDCG). The calculation of \textit{nDCG@k} is given below:
\begin{equation}
    nDCG@k= \frac{DCG_k}{IDCG_k}, 
\end{equation}
\begin{equation}
    DCG_k=\sum_{j=1}^{k} \frac{r_j}{log_2(j+1)},
\end{equation}
where $r_j$ denotes the relevance value at rank position $j$. In the context of location prediction, $r_j \in \{0, 1\}$, and $r_j=1$ if and only if the $j$-th item in the list of predicted locations ranked according to $P(\hat{l}_{n+1})$ matches the ground truth next location. In our experiment, we report the average \textit{nDCG@10} over all test samples.



\section{Results}
\label{sec:res}
\setlength{\tabcolsep}{1pt}  
\begin{table*}[]
\caption{Performance comparison of different embedding methods on next location prediction task. The best and second-best performance are marked in \textbf{bold} and \underline{underlined}, respectively. For better readability, all metric values are scaled by a factor of $10^2$. The \textit{relative difference} (Rel. diff) is also reported, calculated as the \textcolor{deepgreen}{improvement} or \textcolor{deepred}{decrease} of CaLLiPer’s performance relative to the best competing method for each metric.}
\label{tab:perf_res}
\begin{tabular}{cl|ccccc|ccccc}
\toprule
\multicolumn{2}{l}{} & \multicolumn{5}{c}{Conventional} & \multicolumn{5}{c}{Inductive} \\
\midrule
 Data & Embedding & {Acc@1} & {Acc@5} & {Acc@10} & {MRR} & {nDCG@10} & {Acc@1} & {Acc@5} & {Acc@10} & {MRR} & {nDCG@10} \\
\midrule
\multirow{8}{*}{\rotatebox{90}{FSQ-NYC}} & Vanilla-E2E & 19.94±0.19 & \underline{46.84±0.44} & \underline{57.32±0.54} & \underline{32.05±0.18} & \underline{37.52±0.25} & \underline{15.25±1.04} & \underline{35.73±1.66} & \underline{42.44±2.46} & \underline{24.25±1.37} & \underline{28.27±1.56} \\
 & Skip-gram & 19.81±0.34 & 45.76±0.19 & 56.15±0.35 & 31.61±0.29 & 36.85±0.29 & 14.18±1.18 & 33.53±2.50 & 40.25±3.36 & 22.84±1.76 & 26.61±2.11 \\
 & POI2Vec & 19.62±0.58 & 44.94±0.71 & 55.36±0.90& 31.15±0.62 & 36.32±0.66 & 14.68±1.29 & 33.43±3.07 & 40.02±3.88 & 23.10±2.00 & 26.73±2.42 \\
 & Geo-Teaser & 19.43±0.73 & 45.46±0.84 & 55.76±0.48 & 31.24±0.65 & 36.48±0.59 & 13.63±1.00 & 32.20±2.34 & 38.77±3.37 & 22.01±1.62 & 25.62±1.97 \\
 & TALE & \underline{20.20±0.40} & 45.80±0.12 & 55.95±0.27 & 31.80±0.19 & 36.98±0.12 & 14.45±1.10 & 32.96±2.83 & 39.42±3.57 & 22.73±1.82 & 26.36±2.20 \\
 & CTLE & 18.18±0.74 & 44.86±0.37 & 56.58±0.36 & 30.47±0.69 & 35.99±0.59 & 14.17±2.10 & 34.31±5.89 & 41.69±7.59 & 23.07±3.63 & 27.14±4.56 \\
 & CaLLiPer & \textbf{20.33±0.36} & \textbf{48.38±0.15} & \textbf{59.38±0.10} & \textbf{32.97±0.17} & \textbf{38.67±0.12} & \textbf{16.07±0.69} & \textbf{37.68±1.99} & \textbf{45.19±2.76} & \textbf{25.64±1.29} & \textbf{29.94±1.61} \\
 & \multicolumn{1}{l}{\hspace{0.1em} Rel. diff (\%)}  & \textcolor{deepgreen}{+0.64} & \textcolor{deepgreen}{+3.29} & \textcolor{deepgreen}{+3.59} & \textcolor{deepgreen}{+2.87} & \multicolumn{1}{c}{\textcolor{deepgreen}{+3.07}} & \textcolor{deepgreen}{+5.38} & \textcolor{deepgreen}{+5.46} & \textcolor{deepgreen}{+6.48} & \textcolor{deepgreen}{+5.73} & \textcolor{deepgreen}{+5.91} \\
 \midrule
\multirow{8}{*}{\rotatebox{90}{FSQ-TKY}} & Vanilla-E2E & 21.54±0.06 & 45.54±0.16 & 55.39±0.09 & 32.60±0.04 & 37.32±0.03 & 15.75±0.83 & 35.10±1.56 & 42.49±2.09 & 24.62±1.19 & 28.33±1.38 \\
 & Skip-gram & 21.85±0.23 & \underline{46.38±0.10} & 56.32±0.06 & 33.13±0.16 & \underline{37.93±0.13} & \underline{16.22±0.82} & \underline{35.88±1.57} & \underline{43.52±2.02} & 25.28±1.19 & \underline{29.03±1.37} \\
 & POI2Vec & \textbf{22.03±0.13} & 46.35±0.07 & 56.20±0.06 & \underline{33.23±0.08} & 37.98±0.06 & \textbf{16.39±0.97} & 35.78±1.90 & 43.32±2.28 & \underline{25.32±1.36} & 29.03±1.56 \\
 & Geo-Teaser & \underline{21.86±0.19} & \textbf{46.70±0.14} & \textbf{56.73±0.13} & \textbf{33.25±0.16} & \textbf{38.13±0.14} & 16.12±1.12 & 35.81±1.87 & 43.46±2.12 & 25.20±1.45 & 28.97±1.59 \\
 & TALE & 21.44±0.09  & 45.83±0.03 & 55.78±0.13 & 32.68±0.07 & 37.46±0.06 & 16.13±0.82 & 35.82±1.61 & 43.31±2.09 & 25.13±1.18 & 28.91±1.36 \\
 & CTLE & 18.83±0.22 & 44.63±0.13 & 55.00±0.08 & 30.66±0.16 & 35.73±0.14 & 14.12±0.52 & 33.60±1.81 & 41.20±2.45 & 23.09±1.11 & 26.83±1.39 \\
 & CaLLiPer & 20.20±0.14 & 46.19±0.09 & \underline{56.56±0.04} & 32.08±0.09 & 37.19±0.07 & 16.00±0.68 & \textbf{36.46±1.62} & \textbf{44.41±2.13} & \textbf{25.41±1.10} & \textbf{29.34±1.32} \\
 & \multicolumn{1}{l}{\hspace{0.1em} Rel. diff (\%)} & \textcolor{deepred}{-8.31} & \textcolor{deepred}{-1.09} & \textcolor{deepred}{-0.30} & \textcolor{deepred}{-3.52} & \textcolor{deepred}{-2.47} & \textcolor{deepred}{-2.38} & \textcolor{deepgreen}{+1.62} & \textcolor{deepgreen}{+2.05} & \textcolor{deepgreen}{+0.05} & \textcolor{deepgreen}{+1.07} \\
 \midrule
\multirow{8}{*}{\rotatebox{90}{Gowalla-LD}} & Vanilla-E2E & 13.19±0.98 & 28.05±2.14 & 32.29±2.46 & 20.29±1.16 & 22.54±1.43 & 8.28±1.22 & 16.12±0.90 & 19.98±1.67 & 12.37±1.04 & 13.64±1.06 \\
 & Skip-gram & \underline{16.67±0.77} & \underline{35.67±1.36} & \underline{42.24±0.95} & \underline{25.45±0.69} & \underline{28.85±0.70} & \underline{12.21±1.16} & \underline{24.50±2.34} & \underline{28.96±2.01} & \underline{17.92±1.00} & \underline{20.04±1.06} \\
 & POI2Vec & 15.67±0.89  & 33.06±1.49 & 39.85±1.65 & 23.94±1.12 & 27.09±1.26 & 11.78±1.09 & 22.92±0.83 & 27.71±1.17 & 17.36±0.95 & 19.27±1.01 \\
 & Geo-Teaser & 16.48±0.61 & 35.20±0.44 & 41.41±0.70 & 25.12±0.41 & 28.38±0.35 & 11.74±0.79 & 23.29±2.28 & 28.18±2.30 & 17.39±1.22 & 19.43±1.51 \\
 & TALE & 14.73±0.97 & 30.33±1.65 & 35.47±1.89 & 22.09±0.82 & 24.67±0.95 & 10.54±0.71 & 21.41±0.76 & 25.21±0.66 & 15.66±0.66 & 17.44±0.59 \\
 & CTLE & 14.26±2.36 & 32.64±6.33 & 38.93±6.95 & 22.96±3.75 & 26.21±4.61 & 9.67±3.76 & 20.41±8.31 & 24.97±9.36 & 14.94±5.57 & 16.83±6.57 \\
 & CaLLiPer & \textbf{19.06±1.53}  & \textbf{39.36±0.72} & \textbf{46.94±0.87} & \textbf{28.56±0.91} & \textbf{32.41±0.72} & \textbf{13.58±2.00} & \textbf{28.60±3.94} & \textbf{33.79±4.69} & \textbf{20.52±2.69} & \textbf{23.26±3.19} \\
 & \multicolumn{1}{l}{\hspace{0.1em} Rel. diff (\%)} & \textcolor{deepgreen}{+14.34} & \textcolor{deepgreen}{+10.34} & \textcolor{deepgreen}{+11.13} & \textcolor{deepgreen}{+12.22} & \textcolor{deepgreen}{+12.34} & \textcolor{deepgreen}{+11.22} & \textcolor{deepgreen}{+16.73} & \textcolor{deepgreen}{+16.68} & \textcolor{deepgreen}{+14.51} & \textcolor{deepgreen}{+16.07} \\
 \midrule
\multirow{8}{*}{\rotatebox{90}{Geolife}} & Vanilla-E2E & 39.63±1.57 & 64.36±1.09 & 68.89±1.03 & 50.55±1.10 & 54.75±0.97 & 34.42±1.92 & 56.59±4.40 & 61.39±4.80 & 44.16±3.16 & 48.05±3.53 \\
 & Skip-gram & 38.77±1.38 & \underline{64.78±3.06} & \underline{69.13±2.37} & 50.40±1.72 & 54.69±1.83 & 36.35±1.51 & \textbf{59.44±4.33} & \textbf{62.76±5.06} & 46.33±2.70 & 50.08±3.28 \\
 & POI2Vec & 39.29±1.10  & 63.76±1.12 & 68.12±0.49 & 50.14±1.12 & 54.22±0.94 & 35.29±1.06 & 59.22±2.15 & 61.43±3.06 & 46.34±1.36 & 50.49±1.72 \\
 & Geo-Teaser & \underline{41.43±1.22} & 64.49±1.08 & 67.86±0.52 & \underline{51.37±1.04} & \underline{55.19±0.91} & 38.00±2.62 & \underline{59.25±4.81} & \underline{62.66±5.42} & \underline{47.15±3.75} & \underline{50.69±4.13} \\
 & TALE & 40.17±1.38 & \textbf{65.23±1.19} & \textbf{69.24±1.04} & 50.97±1.12 & 55.16±1.07 & \underline{38.21±1.37} & 58.92±2.89 & 61.78±2.37 & 47.14±1.72 & 50.54±1.87 \\
 & CTLE & 38.49±1.00 & 63.67±0.25 & 67.33±0.42 & 49.48±0.42 & 53.54±0.27 & 26.85±1.14 & 59.12±2.93 & 62.21±3.56 & 40.47±1.81 & 46.36±2.28 \\
 & CaLLiPer & \textbf{41.67±1.64} & 64.76±1.88 & 68.63±2.17 & \textbf{51.75±1.30} & \textbf{55.62±1.46} & \textbf{39.22±1.80} & 58.90±3.21 & 62.07±3.69 & \textbf{47.65±2.28} & \textbf{50.96±2.59} \\
 & \multicolumn{1}{l}{\hspace{0.1em} Rel. diff (\%)} & \textcolor{deepgreen}{+0.58} & \textcolor{deepred}{-0.72} & \textcolor{deepred}{-0.88} & \textcolor{deepgreen}{+0.74} & \textcolor{deepgreen}{+0.78} & \textcolor{deepgreen}{+2.46} & \textcolor{deepred}{-0.91} & \textcolor{deepred}{-1.10} & \textcolor{deepgreen}{+1.06} & \textcolor{deepgreen}{+0.53} \\
 \bottomrule
\end{tabular}
\end{table*}

\subsection{Model Performance Comparison}

The predictive performance of all considered methods is presented in Table \ref{tab:perf_res}. In two out of the four datasets, i.e., FSQ-NYC and Gowalla-LD, CaLLiPer significantly outperforms competitive baseline methods across all evaluation metrics under both conventional and inductive settings. On the FSQ-TKY dataset, although CaLLiPer is less competitive under the conventional setting, it achieves the best performance on most metrics (with Acc@1 as the only exception) under the inductive setting. For the Geolife dataset, CaLLiPer consistently achieves the best performance in terms of top-1 accuracy, MRR, and nDCG@10 (three out of the five evaluation metrics) under both conventional and inductive settings.

Overall, CaLLiPer demonstrates the \textbf{most robust performance}, producing \textbf{the best results in the majority of cases}. In contrast, baseline models only occasionally achieve the best performance—for example, on one dataset but not another, under one experimental setting but not the other, or in terms of some evaluation metrics but not others.

Moreover, CaLLiPer shows a particularly notable advantage under the inductive setting. These empirical results demonstrate the potential of applying inductive spatial-semantic location embeddings to improve individual mobility prediction, particularly in scenarios involving previously unseen locations.  

\subsection{Visualising the Learned Location Embeddings}

\begin{figure*}[h]
  \centering
  \includegraphics[width=0.99\linewidth]{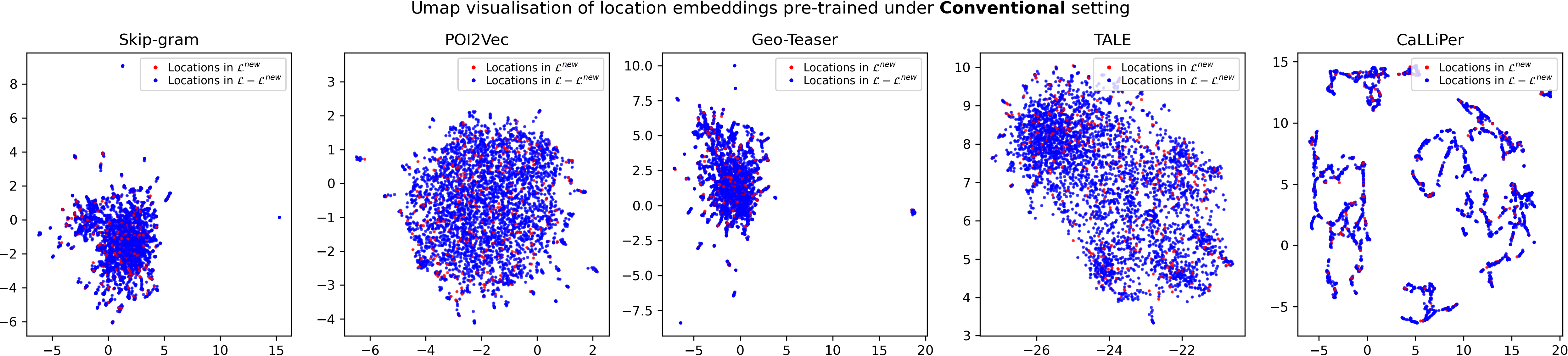}
  \caption{2D visualisation of the embedding vectors pre-trained under \textit{Conventional} setting (using the FSQ-NYC dataset). Note that all locations in the conventinoal setting experiment were available during pre-training; they are coloured differently solely for comparison with the inductive setting results.}
  \label{fig:conv_embed_vis}
\end{figure*}

\begin{figure*}[h]
  \centering
  \includegraphics[width=0.99\linewidth]{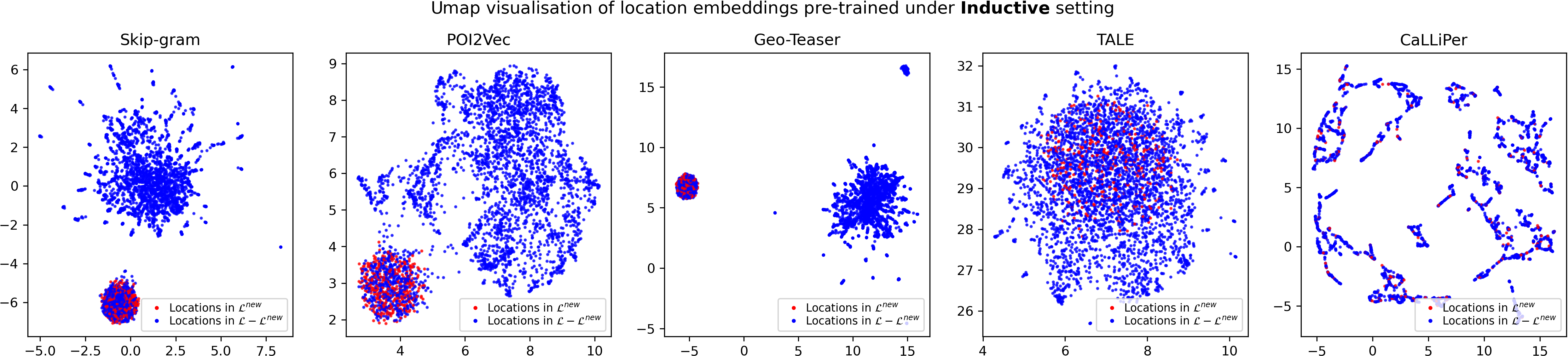}
  \caption{2D visualisation of the embedding vectors pre-trained under \textit{Inductive} setting (using the FSQ-NYC dataset).}
  \label{fig:induct_embed_vis}
\end{figure*}

In this subsection, we visualise the learned embedding vectors of locations from the FSQ-NYC dataset in a two-dimensional space using Uniform Manifold Approximation and
Projection (UMAP) \cite{mcinnes2018umap} for non-linear dimensionality reduction. Figures \ref{fig:conv_embed_vis} and \ref{fig:induct_embed_vis} depict the location embeddings under the \textit{conventional} and \textit{inductive} settings, respectively. These visualisations provide an intuitive view of the relationships among locations in the latent embedding space. For the inductive setting, we distinguish between two types of locations  using different colours: red for locations in the sampled subset $\mathcal{L}^{new}$ (those not seen during pre-training) and blue for locations in $\mathcal{L} - \mathcal{L}^{new}$ (those available during pre-training). Note that all locations in the conventinoal setting experiment were available during pre-training; they are coloured differently solely for comparison with the inductive setting results.

 Figure \ref{fig:conv_embed_vis} shows the embeddings under the conventional scenario. As anticipated, the embedding vectors are well-mixed in the latent space across all methods, indicating that the models have effectively learned unified representation distributions over the entire set of locations.

Comparing the visualisation results in Figure \ref{fig:induct_embed_vis} with that in Figure \ref{fig:conv_embed_vis} helps reveal the model’s ability to generalise to previously unseen locations, i.e., locations in $\mathcal{L}^{new}$. It is evident that, for baseline methods Skip-gram, POI2Vec, and Geo-Teaser, there is a pronounced shift in the distribution of embeddings between the conventional and inductive settings in the latent representation space. Under the inductive setting, the embeddings of new locations generated by these models deviate significantly from those seen during the pre-training. This suggests that the emergence of new locations greatly disrupts the learned manifolds \cite{bengio2013representation}, which is likely to have adverse impacts on the downstream prediction performance. As for the baseline method TALE, it shows a smaller discrepancy between embeddings of new and existing locations under the inductive setting. This improved result might stem from the time-aware objective, which appears to mitigate the divergence between new and seen locations to some extent.

In contrast, the location embeddings learned using CaLLiPer demonstrate great consistency across both settings. The manifolds formed by new and existing locations are well-aligned, indicating CaLLiPer’s strong generalisation capabilities. Notably, the manifolds learned by CaLLiPer have a more fine-grained and clustered structure than those of other baseline methods. This unique pattern might be the result of the location encoding technique employed by CaLLiPer, and similar structures can be found in other works utilising location encoding techniques \cite{klemmer2025satclip}.

\section{Discussion}
\label{sec:discussion}
Learning effective location representations is foundational to individual-level mobility modelling. Conceptually, locations are the basic spatial units where human activity occurs. Methodologically, their vector representations serve as the input to prediction models. Existing methods and our approach tackle this problem from different perspectives. Next, we discuss their core differences, which helps us understand what truly matters in representing locations.

Existing location embedding methods typically learn representations from human mobility sequences. These approaches rely on co-occurrence patterns captured by word2vec-based architectures. The rationale is that the distributed representations can capture complex relationships between locations. Some studies further suggest that characteristics such as the location function can be implicitly learned from mobility patterns \cite{lin2021pre}. 

In contrast, CaLLiPer adopts a task-agnostic approach, learning location embeddings by contrastively aligning spatial coordinates with textual descriptions derived from POI data. Rather than relying on mobility traces, it learns from the inherent structural attributes of urban spaces. CaLLiPer has already demonstrated effectiveness in capturing land use and socio-demographic characteristics \cite{wang2025multi}. This study further shows that CaLLiPer-generated embeddings are also highly effective for individual mobility prediction, especially under the inductive setting, where locations in the test set are unseen during training.

These fundamental differences result in varied performances in downstream location prediction tasks, which makes us ponder what makes a good location representation? What kind of information should be encoded in location vectors? Our findings suggest that for next location prediction, the most valuable signals are not solely indirect co-occurrence patterns, but the inherent characteristics that define a location—namely, its spatial coordinates and semantic (platial) attributes.

Thinking more deeply, location representations can be viewed as part of the “infrastructure” of mobility modelling. Like infrastructure, they should be general-purpose and robust—not tailored to specific tasks or limited user populations. Instead of being trained on narrow objectives derived from mobility traces, they should be grounded in general geographical and semantic features. With such an infrastructure in place, downstream models can specialise in handling specific applications, such as next location prediction, synthetic mobility data generation, or even population flow modelling.

Finally, as access to detailed movement data becomes increasingly restricted due to a growing emphasis on location privacy, the use of publicly available POI data that are not restricted by privacy regulations becomes even more justified as a foundation for learning location representations.

\section{Conclusions}
\label{sec:conclusion}
Current location embedding methods, a cornerstone of human mobility modelling, face issues such as insufficient semantic information modelling and poor generalisation to unseen locations. To address these issues, we propose a timely application of a recent location representation learning method, CaLLiPer, to serve as a location embedder in individuals’ next location prediction. We conducted comprehensive experiments on four public individuals’ mobility datasets. In particular, apart from the conventional experimental setting where the locations during testing are also seen during training, we devised an inductive experimental setting where some locations in the testing set are not seen during training. This simulates the real-world scenario when new locations are visited by individuals. Both the quantitative and qualitative results proved that our application of CaLLiPer outperforms existing location embedding methods. 

For future work, we plan to expand the application of CaLLiPer or its variants \cite{liu2025enriching} in two directions. First, we aim to explore the integration of spatial-semantic embeddings into foundation models, such as large language models (LLMs), to evaluate whether this can enhance LLM-based mobility prediction, which currently represent locations solely as natural language labels \cite{wang2023would}. Second, we intend to scale the application of CaLLiPer beyond the city level to examine its effectiveness across broader geographic contexts.


\bibliographystyle{ACM-Reference-Format}
\bibliography{xlw_refs}


\begin{thebibliography}{41}


\ifx \showCODEN    \undefined \def \showCODEN     #1{\unskip}     \fi
\ifx \showISBNx    \undefined \def \showISBNx     #1{\unskip}     \fi
\ifx \showISBNxiii \undefined \def \showISBNxiii  #1{\unskip}     \fi
\ifx \showISSN     \undefined \def \showISSN      #1{\unskip}     \fi
\ifx \showLCCN     \undefined \def \showLCCN      #1{\unskip}     \fi
\ifx \shownote     \undefined \def \shownote      #1{#1}          \fi
\ifx \showarticletitle \undefined \def \showarticletitle #1{#1}   \fi
\ifx \showURL      \undefined \def \showURL       {\relax}        \fi
\providecommand\bibfield[2]{#2}
\providecommand\bibinfo[2]{#2}
\providecommand\natexlab[1]{#1}
\providecommand\showeprint[2][]{arXiv:#2}

\bibitem[Barbosa et~al\mbox{.}(2018)]%
        {barbosa2018human}
\bibfield{author}{\bibinfo{person}{Hugo Barbosa}, \bibinfo{person}{Marc Barthelemy}, \bibinfo{person}{Gourab Ghoshal}, \bibinfo{person}{Charlotte~R James}, \bibinfo{person}{Maxime Lenormand}, \bibinfo{person}{Thomas Louail}, \bibinfo{person}{Ronaldo Menezes}, \bibinfo{person}{Jos{\'e}~J Ramasco}, \bibinfo{person}{Filippo Simini}, {and} \bibinfo{person}{Marcello Tomasini}.} \bibinfo{year}{2018}\natexlab{}.
\newblock \showarticletitle{Human mobility: Models and applications}.
\newblock \bibinfo{journal}{\emph{Physics Reports}}  \bibinfo{volume}{734} (\bibinfo{year}{2018}), \bibinfo{pages}{1--74}.
\newblock


\bibitem[Bengio et~al\mbox{.}(2013)]%
        {bengio2013representation}
\bibfield{author}{\bibinfo{person}{Yoshua Bengio}, \bibinfo{person}{Aaron Courville}, {and} \bibinfo{person}{Pascal Vincent}.} \bibinfo{year}{2013}\natexlab{}.
\newblock \showarticletitle{Representation learning: A review and new perspectives}.
\newblock \bibinfo{journal}{\emph{IEEE transactions on pattern analysis and machine intelligence}} \bibinfo{volume}{35}, \bibinfo{number}{8} (\bibinfo{year}{2013}), \bibinfo{pages}{1798--1828}.
\newblock


\bibitem[Buchin et~al\mbox{.}(2012)]%
        {buchin2012context}
\bibfield{author}{\bibinfo{person}{Maike Buchin}, \bibinfo{person}{Somayeh Dodge}, {and} \bibinfo{person}{Bettina Speckmann}.} \bibinfo{year}{2012}\natexlab{}.
\newblock \showarticletitle{Context-aware similarity of trajectories}. In \bibinfo{booktitle}{\emph{Geographic Information Science: 7th International Conference, GIScience 2012, Columbus, OH, USA, September 18-21, 2012. Proceedings 7}}. Springer, \bibinfo{pages}{43--56}.
\newblock


\bibitem[Cho et~al\mbox{.}(2011)]%
        {cho2011friendship}
\bibfield{author}{\bibinfo{person}{Eunjoon Cho}, \bibinfo{person}{Seth~A Myers}, {and} \bibinfo{person}{Jure Leskovec}.} \bibinfo{year}{2011}\natexlab{}.
\newblock \showarticletitle{Friendship and mobility: user movement in location-based social networks}. In \bibinfo{booktitle}{\emph{Proceedings of the 17th ACM SIGKDD international conference on Knowledge discovery and data mining}}. \bibinfo{pages}{1082--1090}.
\newblock


\bibitem[Chu et~al\mbox{.}(2019)]%
        {chu2019geo}
\bibfield{author}{\bibinfo{person}{Grace Chu}, \bibinfo{person}{Brian Potetz}, \bibinfo{person}{Weijun Wang}, \bibinfo{person}{Andrew Howard}, \bibinfo{person}{Yang Song}, \bibinfo{person}{Fernando Brucher}, \bibinfo{person}{Thomas Leung}, {and} \bibinfo{person}{Hartwig Adam}.} \bibinfo{year}{2019}\natexlab{}.
\newblock \showarticletitle{Geo-aware networks for fine-grained recognition}. In \bibinfo{booktitle}{\emph{Proceedings of the IEEE/CVF International Conference on Computer Vision Workshops}}.
\newblock


\bibitem[Devlin et~al\mbox{.}(2019)]%
        {devlin2019bert}
\bibfield{author}{\bibinfo{person}{Jacob Devlin}, \bibinfo{person}{Chang Ming-Wei}, \bibinfo{person}{Lee Kenton}, {and} \bibinfo{person}{Toutanova Kristina}.} \bibinfo{year}{2019}\natexlab{}.
\newblock \showarticletitle{Bert: Pre-training of deep bidirectional transformers for language understanding}. In \bibinfo{booktitle}{\emph{Proceedings of naacL-HLT}}, Vol.~\bibinfo{volume}{1}. Minneapolis, Minnesota, \bibinfo{pages}{2}.
\newblock


\bibitem[Fan et~al\mbox{.}(2025)]%
        {fan2025coverage}
\bibfield{author}{\bibinfo{person}{Zicheng Fan}, \bibinfo{person}{Chen-Chieh Feng}, {and} \bibinfo{person}{Filip Biljecki}.} \bibinfo{year}{2025}\natexlab{}.
\newblock \showarticletitle{Coverage and bias of street view imagery in mapping the urban environment}.
\newblock \bibinfo{journal}{\emph{Computers, Environment and Urban Systems}}  \bibinfo{volume}{117} (\bibinfo{year}{2025}), \bibinfo{pages}{102253}.
\newblock


\bibitem[Feng et~al\mbox{.}(2017)]%
        {feng2017poi2vec}
\bibfield{author}{\bibinfo{person}{Shanshan Feng}, \bibinfo{person}{Gao Cong}, \bibinfo{person}{Bo An}, {and} \bibinfo{person}{Yeow~Meng Chee}.} \bibinfo{year}{2017}\natexlab{}.
\newblock \showarticletitle{Poi2vec: Geographical latent representation for predicting future visitors}. In \bibinfo{booktitle}{\emph{Proceedings of the AAAI Conference on Artificial Intelligence}}, Vol.~\bibinfo{volume}{31}.
\newblock


\bibitem[Foursquare(2024)]%
        {fsq2024osp}
\bibfield{author}{\bibinfo{person}{Foursquare}.} \bibinfo{year}{2024}\natexlab{}.
\newblock \bibinfo{booktitle}{\emph{Foursquare Open Source Places: A new foundational dataset for the geospatial community}}.
\newblock
\urldef\tempurl%
\url{https://location.foursquare.com/resources/blog/products/foursquare-open-source-places-a-new-foundational-dataset-for-the-geospatial-community/}
\showURL{%
Retrieved May 31, 2025 from \tempurl}


\bibitem[Hochreiter and Schmidhuber(1997)]%
        {hochreiter1997long}
\bibfield{author}{\bibinfo{person}{Sepp Hochreiter} {and} \bibinfo{person}{J{\"u}rgen Schmidhuber}.} \bibinfo{year}{1997}\natexlab{}.
\newblock \showarticletitle{Long short-term memory}.
\newblock \bibinfo{journal}{\emph{Neural computation}} \bibinfo{volume}{9}, \bibinfo{number}{8} (\bibinfo{year}{1997}), \bibinfo{pages}{1735--1780}.
\newblock


\bibitem[Hong et~al\mbox{.}(2023)]%
        {hong2023context}
\bibfield{author}{\bibinfo{person}{Ye Hong}, \bibinfo{person}{Yatao Zhang}, \bibinfo{person}{Konrad Schindler}, {and} \bibinfo{person}{Martin Raubal}.} \bibinfo{year}{2023}\natexlab{}.
\newblock \showarticletitle{Context-aware multi-head self-attentional neural network model for next location prediction}.
\newblock \bibinfo{journal}{\emph{Transportation Research Part C: Emerging Technologies}}  \bibinfo{volume}{156} (\bibinfo{year}{2023}), \bibinfo{pages}{104315}.
\newblock


\bibitem[Ilyankou et~al\mbox{.}(2023)]%
        {ilyankou2023supermarket}
\bibfield{author}{\bibinfo{person}{Ilya Ilyankou}, \bibinfo{person}{Andy Newing}, {and} \bibinfo{person}{Nick Hood}.} \bibinfo{year}{2023}\natexlab{}.
\newblock \showarticletitle{Supermarket store locations as a proxy for neighbourhood health, wellbeing, and wealth}.
\newblock \bibinfo{journal}{\emph{Sustainability}} \bibinfo{volume}{15}, \bibinfo{number}{15} (\bibinfo{year}{2023}), \bibinfo{pages}{11641}.
\newblock


\bibitem[Klemmer et~al\mbox{.}(2025)]%
        {klemmer2025satclip}
\bibfield{author}{\bibinfo{person}{Konstantin Klemmer}, \bibinfo{person}{Esther Rolf}, \bibinfo{person}{Caleb Robinson}, \bibinfo{person}{Lester Mackey}, {and} \bibinfo{person}{Marc Ru{\ss}wurm}.} \bibinfo{year}{2025}\natexlab{}.
\newblock \showarticletitle{Satclip: Global, general-purpose location embeddings with satellite imagery}. In \bibinfo{booktitle}{\emph{Proceedings of the AAAI Conference on Artificial Intelligence}}, Vol.~\bibinfo{volume}{39}. \bibinfo{pages}{4347--4355}.
\newblock


\bibitem[Kong and Wu(2018)]%
        {kong2018hst}
\bibfield{author}{\bibinfo{person}{Dejiang Kong} {and} \bibinfo{person}{Fei Wu}.} \bibinfo{year}{2018}\natexlab{}.
\newblock \showarticletitle{HST-LSTM: A hierarchical spatial-temporal long-short term memory network for location prediction.}. In \bibinfo{booktitle}{\emph{Ijcai}}, Vol.~\bibinfo{volume}{18}. \bibinfo{pages}{2341--2347}.
\newblock


\bibitem[Lee and Holme(2015)]%
        {lee2015relating}
\bibfield{author}{\bibinfo{person}{Minjin Lee} {and} \bibinfo{person}{Petter Holme}.} \bibinfo{year}{2015}\natexlab{}.
\newblock \showarticletitle{Relating land use and human intra-city mobility}.
\newblock \bibinfo{journal}{\emph{PloS one}} \bibinfo{volume}{10}, \bibinfo{number}{10} (\bibinfo{year}{2015}), \bibinfo{pages}{e0140152}.
\newblock


\bibitem[Lin et~al\mbox{.}(2021)]%
        {lin2021pre}
\bibfield{author}{\bibinfo{person}{Yan Lin}, \bibinfo{person}{Huaiyu Wan}, \bibinfo{person}{Shengnan Guo}, {and} \bibinfo{person}{Youfang Lin}.} \bibinfo{year}{2021}\natexlab{}.
\newblock \showarticletitle{Pre-training context and time aware location embeddings from spatial-temporal trajectories for user next location prediction}. In \bibinfo{booktitle}{\emph{Proceedings of the AAAI Conference on Artificial Intelligence}}, Vol.~\bibinfo{volume}{35}. \bibinfo{pages}{4241--4248}.
\newblock


\bibitem[Liu et~al\mbox{.}(2025)]%
        {liu2025enriching}
\bibfield{author}{\bibinfo{person}{Junyuan Liu}, \bibinfo{person}{Xinglei Wang}, {and} \bibinfo{person}{Tao Cheng}.} \bibinfo{year}{2025}\natexlab{}.
\newblock \showarticletitle{Enriching Location Representation with Detailed Semantic Information}.
\newblock \bibinfo{journal}{\emph{arXiv preprint arXiv:2506.02744}} (\bibinfo{year}{2025}).
\newblock


\bibitem[Liu et~al\mbox{.}(2016)]%
        {liu2016exploring}
\bibfield{author}{\bibinfo{person}{Xin Liu}, \bibinfo{person}{Yong Liu}, {and} \bibinfo{person}{Xiaoli Li}.} \bibinfo{year}{2016}\natexlab{}.
\newblock \showarticletitle{Exploring the context of locations for personalized location recommendations.}. In \bibinfo{booktitle}{\emph{IJCAI}}. \bibinfo{pages}{1188--1194}.
\newblock


\bibitem[Luca et~al\mbox{.}(2021)]%
        {luca2021survey}
\bibfield{author}{\bibinfo{person}{Massimiliano Luca}, \bibinfo{person}{Gianni Barlacchi}, \bibinfo{person}{Bruno Lepri}, {and} \bibinfo{person}{Luca Pappalardo}.} \bibinfo{year}{2021}\natexlab{}.
\newblock \showarticletitle{A survey on deep learning for human mobility}.
\newblock \bibinfo{journal}{\emph{ACM Computing Surveys (CSUR)}} \bibinfo{volume}{55}, \bibinfo{number}{1} (\bibinfo{year}{2021}), \bibinfo{pages}{1--44}.
\newblock


\bibitem[Mac~Aodha et~al\mbox{.}(2019)]%
        {mac2019presence}
\bibfield{author}{\bibinfo{person}{Oisin Mac~Aodha}, \bibinfo{person}{Elijah Cole}, {and} \bibinfo{person}{Pietro Perona}.} \bibinfo{year}{2019}\natexlab{}.
\newblock \showarticletitle{Presence-only geographical priors for fine-grained image classification}. In \bibinfo{booktitle}{\emph{Proceedings of the IEEE/CVF International Conference on Computer Vision}}. \bibinfo{pages}{9596--9606}.
\newblock


\bibitem[Mai et~al\mbox{.}(2022)]%
        {mai2022review}
\bibfield{author}{\bibinfo{person}{Gengchen Mai}, \bibinfo{person}{Krzysztof Janowicz}, \bibinfo{person}{Yingjie Hu}, \bibinfo{person}{Song Gao}, \bibinfo{person}{Bo Yan}, \bibinfo{person}{Rui Zhu}, \bibinfo{person}{Ling Cai}, {and} \bibinfo{person}{Ni Lao}.} \bibinfo{year}{2022}\natexlab{}.
\newblock \showarticletitle{A review of location encoding for GeoAI: methods and applications}.
\newblock \bibinfo{journal}{\emph{International Journal of Geographical Information Science}}  \bibinfo{volume}{36} (\bibinfo{year}{2022}), \bibinfo{pages}{639--673}.
\newblock
\showISSN{1365-8816, 1362-3087}
\href{https://doi.org/10.1080/13658816.2021.2004602}{doi:\nolinkurl{10.1080/13658816.2021.2004602}}


\bibitem[Mai et~al\mbox{.}(2020)]%
        {mai2020iclr}
\bibfield{author}{\bibinfo{person}{Gengchen Mai}, \bibinfo{person}{Krzysztof Janowicz}, \bibinfo{person}{Bo Yan}, \bibinfo{person}{Rui Zhu}, \bibinfo{person}{Ling Cai}, {and} \bibinfo{person}{Ni Lao}.} \bibinfo{year}{2020}\natexlab{}.
\newblock \showarticletitle{Multi-Scale Representation Learning for Spatial Feature Distributions using Grid Cells}. In \bibinfo{booktitle}{\emph{International Conference on Learning Representations}}.
\newblock
\urldef\tempurl%
\url{https://openreview.net/forum?id=rJljdh4KDH}
\showURL{%
\tempurl}


\bibitem[Mai et~al\mbox{.}(2023)]%
        {mai2023sphere2vec}
\bibfield{author}{\bibinfo{person}{Gengchen Mai}, \bibinfo{person}{Yao Xuan}, \bibinfo{person}{Wenyun Zuo}, \bibinfo{person}{Yutong He}, \bibinfo{person}{Jiaming Song}, \bibinfo{person}{Stefano Ermon}, \bibinfo{person}{Krzysztof Janowicz}, {and} \bibinfo{person}{Ni Lao}.} \bibinfo{year}{2023}\natexlab{}.
\newblock \showarticletitle{Sphere2Vec: A general-purpose location representation learning over a spherical surface for large-scale geospatial predictions}.
\newblock \bibinfo{journal}{\emph{ISPRS Journal of Photogrammetry and Remote Sensing}}  \bibinfo{volume}{202} (\bibinfo{year}{2023}), \bibinfo{pages}{439--462}.
\newblock


\bibitem[Martin et~al\mbox{.}(2023)]%
        {martin2023trackintel}
\bibfield{author}{\bibinfo{person}{Henry Martin}, \bibinfo{person}{Ye Hong}, \bibinfo{person}{Nina Wiedemann}, \bibinfo{person}{Dominik Bucher}, {and} \bibinfo{person}{Martin Raubal}.} \bibinfo{year}{2023}\natexlab{}.
\newblock \showarticletitle{Trackintel: An open-source Python library for human mobility analysis}.
\newblock \bibinfo{journal}{\emph{Computers, Environment and Urban Systems}}  \bibinfo{volume}{101} (\bibinfo{year}{2023}), \bibinfo{pages}{101938}.
\newblock


\bibitem[McInnes et~al\mbox{.}(2018)]%
        {mcinnes2018umap}
\bibfield{author}{\bibinfo{person}{Leland McInnes}, \bibinfo{person}{John Healy}, {and} \bibinfo{person}{James Melville}.} \bibinfo{year}{2018}\natexlab{}.
\newblock \showarticletitle{Umap: Uniform manifold approximation and projection for dimension reduction}.
\newblock \bibinfo{journal}{\emph{arXiv preprint arXiv:1802.03426}} (\bibinfo{year}{2018}).
\newblock


\bibitem[Mikolov(2013)]%
        {mikolov2013efficient}
\bibfield{author}{\bibinfo{person}{Tomas Mikolov}.} \bibinfo{year}{2013}\natexlab{}.
\newblock \showarticletitle{Efficient estimation of word representations in vector space}.
\newblock \bibinfo{journal}{\emph{arXiv preprint arXiv:1301.3781}} (\bibinfo{year}{2013}).
\newblock


\bibitem[Mikolov et~al\mbox{.}(2013)]%
        {mikolov2013distributed}
\bibfield{author}{\bibinfo{person}{Tomas Mikolov}, \bibinfo{person}{Ilya Sutskever}, \bibinfo{person}{Kai Chen}, \bibinfo{person}{Greg~S Corrado}, {and} \bibinfo{person}{Jeff Dean}.} \bibinfo{year}{2013}\natexlab{}.
\newblock \showarticletitle{Distributed representations of words and phrases and their compositionality}.
\newblock \bibinfo{journal}{\emph{Advances in neural information processing systems}}  \bibinfo{volume}{26} (\bibinfo{year}{2013}).
\newblock


\bibitem[Nils and Iryna(2019)]%
        {reimers2019sbert}
\bibfield{author}{\bibinfo{person}{Reimers Nils} {and} \bibinfo{person}{Gurevych Iryna}.} \bibinfo{year}{2019}\natexlab{}.
\newblock \showarticletitle{Sentence-BERT: Sentence Embeddings using Siamese BERT-Networks}. In \bibinfo{booktitle}{\emph{Proceedings of the 2019 Conference on Empirical Methods in Natural Language Processing and the 9th International Joint Conference on Natural Language Processing (EMNLP-IJCNLP)}}. \bibinfo{pages}{3982--3992}.
\newblock


\bibitem[Pappalardo et~al\mbox{.}(2023)]%
        {pappalardo2023future}
\bibfield{author}{\bibinfo{person}{Luca Pappalardo}, \bibinfo{person}{Ed Manley}, \bibinfo{person}{Vedran Sekara}, {and} \bibinfo{person}{Laura Alessandretti}.} \bibinfo{year}{2023}\natexlab{}.
\newblock \showarticletitle{Future directions in human mobility science}.
\newblock \bibinfo{journal}{\emph{Nature computational science}} \bibinfo{volume}{3}, \bibinfo{number}{7} (\bibinfo{year}{2023}), \bibinfo{pages}{588--600}.
\newblock


\bibitem[Radford et~al\mbox{.}(2021)]%
        {radford2021clip}
\bibfield{author}{\bibinfo{person}{Alec Radford}, \bibinfo{person}{Jong~Wook Kim}, \bibinfo{person}{Chris Hallacy}, \bibinfo{person}{Aditya Ramesh}, \bibinfo{person}{Gabriel Goh}, \bibinfo{person}{Sandhini Agarwal}, \bibinfo{person}{Girish Sastry}, \bibinfo{person}{Amanda Askell}, \bibinfo{person}{Pamela Mishkin}, \bibinfo{person}{Jack Clark}, {et~al\mbox{.}}} \bibinfo{year}{2021}\natexlab{}.
\newblock \showarticletitle{Learning transferable visual models from natural language supervision}. In \bibinfo{booktitle}{\emph{International conference on machine learning}}. PMLR, \bibinfo{pages}{8748--8763}.
\newblock


\bibitem[Ru{\ss}wurm et~al\mbox{.}(2024)]%
        {russwurm2024sh}
\bibfield{author}{\bibinfo{person}{Marc Ru{\ss}wurm}, \bibinfo{person}{Konstantin Klemmer}, \bibinfo{person}{Esther Rolf}, \bibinfo{person}{Robin Zbinden}, {and} \bibinfo{person}{Devis Tuia}.} \bibinfo{year}{2024}\natexlab{}.
\newblock \showarticletitle{Geographic Location Encoding with Spherical Harmonics and Sinusoidal Representation Networks}. In \bibinfo{booktitle}{\emph{The 12th International Conference on Learning Representations}}.
\newblock


\bibitem[Vaswani et~al\mbox{.}(2017)]%
        {vaswani2017attention}
\bibfield{author}{\bibinfo{person}{Ashish Vaswani}, \bibinfo{person}{Noam Shazeer}, \bibinfo{person}{Niki Parmar}, \bibinfo{person}{Jakob Uszkoreit}, \bibinfo{person}{Llion Jones}, \bibinfo{person}{Aidan~N Gomez}, \bibinfo{person}{{\L}ukasz Kaiser}, {and} \bibinfo{person}{Illia Polosukhin}.} \bibinfo{year}{2017}\natexlab{}.
\newblock \showarticletitle{Attention is all you need}.
\newblock \bibinfo{journal}{\emph{Advances in neural information processing systems}}  \bibinfo{volume}{30} (\bibinfo{year}{2017}).
\newblock


\bibitem[Wan et~al\mbox{.}(2021)]%
        {wan2021pre}
\bibfield{author}{\bibinfo{person}{Huaiyu Wan}, \bibinfo{person}{Yan Lin}, \bibinfo{person}{Shengnan Guo}, {and} \bibinfo{person}{Youfang Lin}.} \bibinfo{year}{2021}\natexlab{}.
\newblock \showarticletitle{Pre-training time-aware location embeddings from spatial-temporal trajectories}.
\newblock \bibinfo{journal}{\emph{IEEE Transactions on Knowledge and Data Engineering}} \bibinfo{volume}{34}, \bibinfo{number}{11} (\bibinfo{year}{2021}), \bibinfo{pages}{5510--5523}.
\newblock


\bibitem[Wang et~al\mbox{.}(2025)]%
        {wang2025multi}
\bibfield{author}{\bibinfo{person}{Xinglei Wang}, \bibinfo{person}{Tao Cheng}, \bibinfo{person}{Stephen Law}, \bibinfo{person}{Zichao Zeng}, \bibinfo{person}{Lu Yin}, {and} \bibinfo{person}{Junyuan Liu}.} \bibinfo{year}{2025}\natexlab{}.
\newblock \showarticletitle{Multi-modal contrastive learning of urban space representations from POI data}.
\newblock \bibinfo{journal}{\emph{Computers, Environment and Urban Systems}}  \bibinfo{volume}{120} (\bibinfo{year}{2025}), \bibinfo{pages}{102299}.
\newblock


\bibitem[Wang et~al\mbox{.}(2023)]%
        {wang2023would}
\bibfield{author}{\bibinfo{person}{Xinglei Wang}, \bibinfo{person}{Meng Fang}, \bibinfo{person}{Zichao Zeng}, {and} \bibinfo{person}{Tao Cheng}.} \bibinfo{year}{2023}\natexlab{}.
\newblock \showarticletitle{Where would i go next? large language models as human mobility predictors}.
\newblock \bibinfo{journal}{\emph{arXiv preprint arXiv:2308.15197}} (\bibinfo{year}{2023}).
\newblock


\bibitem[Yang et~al\mbox{.}(2014)]%
        {yang2014modeling}
\bibfield{author}{\bibinfo{person}{Dingqi Yang}, \bibinfo{person}{Daqing Zhang}, \bibinfo{person}{Vincent~W Zheng}, {and} \bibinfo{person}{Zhiyong Yu}.} \bibinfo{year}{2014}\natexlab{}.
\newblock \showarticletitle{Modeling user activity preference by leveraging user spatial temporal characteristics in LBSNs}.
\newblock \bibinfo{journal}{\emph{IEEE Transactions on Systems, Man, and Cybernetics: Systems}} \bibinfo{volume}{45}, \bibinfo{number}{1} (\bibinfo{year}{2014}), \bibinfo{pages}{129--142}.
\newblock


\bibitem[Yao et~al\mbox{.}(2018)]%
        {yao2018representing}
\bibfield{author}{\bibinfo{person}{Zijun Yao}, \bibinfo{person}{Yanjie Fu}, \bibinfo{person}{Bin Liu}, \bibinfo{person}{Wangsu Hu}, {and} \bibinfo{person}{Hui Xiong}.} \bibinfo{year}{2018}\natexlab{}.
\newblock \showarticletitle{Representing urban functions through zone embedding with human mobility patterns}. In \bibinfo{booktitle}{\emph{Proceedings of the Twenty-Seventh International Joint Conference on Artificial Intelligence (IJCAI-18)}}.
\newblock


\bibitem[Zhao et~al\mbox{.}(2020)]%
        {zhao2020go}
\bibfield{author}{\bibinfo{person}{Pengpeng Zhao}, \bibinfo{person}{Anjing Luo}, \bibinfo{person}{Yanchi Liu}, \bibinfo{person}{Jiajie Xu}, \bibinfo{person}{Zhixu Li}, \bibinfo{person}{Fuzhen Zhuang}, \bibinfo{person}{Victor~S Sheng}, {and} \bibinfo{person}{Xiaofang Zhou}.} \bibinfo{year}{2020}\natexlab{}.
\newblock \showarticletitle{Where to go next: A spatio-temporal gated network for next poi recommendation}.
\newblock \bibinfo{journal}{\emph{IEEE Transactions on Knowledge and Data Engineering}} \bibinfo{volume}{34}, \bibinfo{number}{5} (\bibinfo{year}{2020}), \bibinfo{pages}{2512--2524}.
\newblock


\bibitem[Zhao et~al\mbox{.}(2017)]%
        {zhao2017geo}
\bibfield{author}{\bibinfo{person}{Shenglin Zhao}, \bibinfo{person}{Tong Zhao}, \bibinfo{person}{Irwin King}, {and} \bibinfo{person}{Michael~R Lyu}.} \bibinfo{year}{2017}\natexlab{}.
\newblock \showarticletitle{Geo-teaser: Geo-temporal sequential embedding rank for point-of-interest recommendation}. In \bibinfo{booktitle}{\emph{Proceedings of the 26th international conference on world wide web companion}}. \bibinfo{pages}{153--162}.
\newblock


\bibitem[Zheng et~al\mbox{.}(2010)]%
        {zheng2010geolife}
\bibfield{author}{\bibinfo{person}{Yu Zheng}, \bibinfo{person}{Xing Xie}, \bibinfo{person}{Wei-Ying Ma}, {et~al\mbox{.}}} \bibinfo{year}{2010}\natexlab{}.
\newblock \showarticletitle{GeoLife: A collaborative social networking service among user, location and trajectory.}
\newblock \bibinfo{journal}{\emph{IEEE Data Eng. Bull.}} \bibinfo{volume}{33}, \bibinfo{number}{2} (\bibinfo{year}{2010}), \bibinfo{pages}{32--39}.
\newblock


\bibitem[Zhou and Huang(2018)]%
        {zhou2018deepmove}
\bibfield{author}{\bibinfo{person}{Yang Zhou} {and} \bibinfo{person}{Yan Huang}.} \bibinfo{year}{2018}\natexlab{}.
\newblock \showarticletitle{Deepmove: Learning place representations through large scale movement data}. In \bibinfo{booktitle}{\emph{2018 IEEE international conference on big data (big data)}}. IEEE, \bibinfo{pages}{2403--2412}.
\newblock


\end{thebibliography}

\appendix

\section{More experimental details}

To ensure a fair comparison, we set the embedding dimension to 128 across all methods, and the architecture of the downstream model is kept consistent for all compared methods. The hyperparameter settings for the downstream model (MHSA \cite{hong2023context}) are as follows: number of layers = 6, number of attention heads = 8, feed-forward layer dimension = 256, and dropout rate = 0.1.

Additional implementation details for CaLLiPer and the baseline methods are as follows:

\subsection{CaLLiPer}
\label{append:calliper}

The implementation of CaLLiPer follows the original paper \cite{wang2025multi}. For the location encoder, we employ Grid \cite{mai2020iclr} as the PE and FC-Net as the NN. The mathematical formulation of Grid is as follows:

\begin{equation}
    \text{PE}(\lambda,\phi)=\bigcup_{s=0}^{S-1}(\cos \frac{\lambda}{\alpha_s} ,\sin \frac{\lambda}{\alpha_{s}},\cos \frac{\phi}{\alpha_s},\sin \frac{\phi}{\alpha_s})
\end{equation}

\begin{equation}
    \alpha_{s}=r_{\text{min}} \cdot (\frac{r_{\text{max}}}{r_{\text{min}}})^{\frac{s}{S-1}}
\end{equation} where $r_{\text{min}}$ and $r_{\text{max}}$ are the minimum and maximum radii, respectively, and $S$ is the number of scales. These hyperparameters control the resolutions of the muti-scale encoding of the coordinates. We adopted Sentence Transformer \cite{reimers2019sbert} as the text encoder $f^T$ and a linear layer as the projection layer $f^P$.

The hyperparameters of Grid vary across different datasets: $r_{\text{min}}$ and $r_{\text{max}}$ are set to 0.01 and 10 for FSQ-NYC and FSQ-TKY, 1 and 1000 for Gowalla-LD, and 0.01 and 10 for Geolife. All other hyperparameters remain consistent: we set the number of scales $S=32$ and the hidden dimension of the FC-Net as 256. 

The training parameters are as follows: learning rate: 0.001; optimiser: Adam. Batch sizes are 128 for FSQ-NYC, 256 for FSQ-TKY, 1024 for Gowalla-LD, and 256 for Geolife.

\subsection{Baseline methods}

For baseline methods, we tuned the hyperparameters via random search and pre-trained them until convergence using a learning rate of 0.001 and the Adam optimiser. As the numerous hyperparameters vary significantly across different methods and datasets, it is impractical to list them all here. Therefore, we refer readers to our open-source code repository at https://github.com/xlwang233/Into-the-Unknown for detailed hyperparameter settings.

\end{document}